# Stochastic Deconvolutional Neural Network Ensemble Training on Generative Pseudo-Adversarial Networks

## A Method for Minimising the Mode Collapse Problem

**Alexey Chaplygin & Joshua Chacksfield**
Data Science Research Group, Department of Analytics and Business Intelligence, PVH Corp.
{alexey.chapligin, chacksfieldj}@gmail.com

### ABSTRACT

*The training of Generative Adversarial Networks (GANs) is a difficult task mainly due to the nature of the networks. One such issue is when the generator and discriminator start oscillating, rather than converging to a fixed point. Another case can be when one agent becomes more adept than the other which results in the decrease of the other agent's ability to learn, reducing the learning capacity of the system as a whole. Additionally, there exists the problem of 'Mode Collapse' which involves the generators output collapsing to a single sample or a small set of similar samples. To train GANs a careful selection of the architecture that is used along with a variety of other methods to improve training. Even when applying these methods there is low stability of training in relation to the parameters that are chosen. Stochastic ensembling is suggested as a method for improving the stability while training GANs.*

## Introduction

Deep Networks have made great advances in the area of generative modes. These advances have been the result of a wide range of training losses and architectures, including but limited to Generative Adversarial Networks (GANs) [1].

Most deep generative models are trained by the use of two models. They are used to solve a minimax 'game', with a Generator $G$ sampling data, Discriminator $D$ classifying the data as real or generated. In theory these models are capable of modeling an arbitrarily complex probability distribution. The ability to train flexible generating functions have made GANs extremely successful in image generation [2].

In practice, however, GANs suffer from many issues, particularly during training. One common failure mode involves the generator collapsing to produce only a single sample or a small family of very similar samples. Another involves the generator and discriminator oscillating during training, rather than converging to a fixed point. In addition, if one agent becomes much more powerful than the other, the learning signal to the other agent becomes useless, and the system does not learn.

A lot of attempts have been made to minimise the mode-collapse problem and improve variety on the output [3, 4, 5, 6]. However, some solutions are computationally expensive and treated mode-collapse problem symptomatically.

The assumption behind the methodology described later is that the architecture of a typical GAN causes mode-collapse to occur. The Discriminator $D$ portion of the network constantly requires new samples from the Generator $G$ and due to how $D$ is defined, it never reaches a state for which the output of $G$ is satisfactory. This in turn results into two possible ways for the model to evolve. Firstly, in the case where $G$ is overly powerful the network can start oscillating. This is where even the slightest modification of parameters can result in significantly different outputs that the discriminator cannot "remember". In this situation the output differs from epoch to epoch, at the cost of local variety inside of one epoch. We call this scenario the "soft-collapse" of a model.

However, if $G$ is weaker the oscillation scenario cannot occur. In this instance a situation called "hard-collapse" may manifest. This is where after a small number of attempts to significantly modify the output and go into oscillation mode, it fails. The discriminator becomes absolute certain that the all samples are fake. This results with the loss of the generator being effectively infinite. This results in undefined gradients and it being impossible for the training to progress further.

As we believe that mode-collapse is unavoidable situation another, synthetic way of solving this issue, is suggested. Proposed is the simple idea of Stochastic Ensembling, which can be described as random shuffling of filters on deep levels of the generator. This is comparable to creating a set of weak generators that can still suffer from the mode-collapse problem, but still produce an acceptable output variety.

The efficiency of the described method is demonstrated on another, Pseudo-GAN, where the role of discriminator is played by any pre-trained image classifier. This can be seen as a state of absolute mode-collapse from the beginning of training.

## Methodology

The main difference between standard GAN architecture and one using *Stochastic Ensembling* is in the way that the deep layers are constructed within the Generator. In these layers, stochastic deconvolution is applied, the main idea of which is to randomly select a set of filters from a fixed filter bank.

In this architecture a *stochastic deconvolution* layer is constructed using filters of size 4, applied with a stride of size 2. PReLU (Parameterised Leaky ReLU function initialised to 0.2) was applied to improve model fitting [7] and weight normalization was also used to improve stability [8].

The higher level layers are left as standard deconvolution layers so as to provide refinement for the network and can be reused between different combinations of deep layers. Meanwhile different combinations of deep layers are available for covering the different distributions in the training dataset. An architecture that could achieve the same effect as Stochastic Ensembling is to split the generator into an ensemble of generators with shared upper layers. This increases the size of the networks requirements making it computationally expensive to train. On the other hand stochastic deconvolutions create $8^4 = 4096$ different 'routes' through the use of only 8 different filters in 4 deep layers.

From an intuitive point of view, the combination of paths covers different visual "topics" in the training distribution, for which high-level features are usually shared. This prevents the network from early collapse and describes the distribution more effectively. It does not guarantee that GANs based on stochastic deconvolution do not suffer from mode collapse, but does provide some redundancy. Even if each route of 4096 in the example above collapsed it would still provide some variety. Another benefit of using stochastic deconvolutional layers is that the size of filters can be kept smaller. This enables the discriminator to outperform each sub-generator and in the worst case scenario the sub-generator will start oscillations without experiencing hard mode collapse.

We believe that stochastic ensembling can be beneficial outside of GAN models and can be useful for any problem that involves generative models. The approach could offer an avenue of further research for the application in non-generative models as an easily implemented alternative to other ensemble techniques.

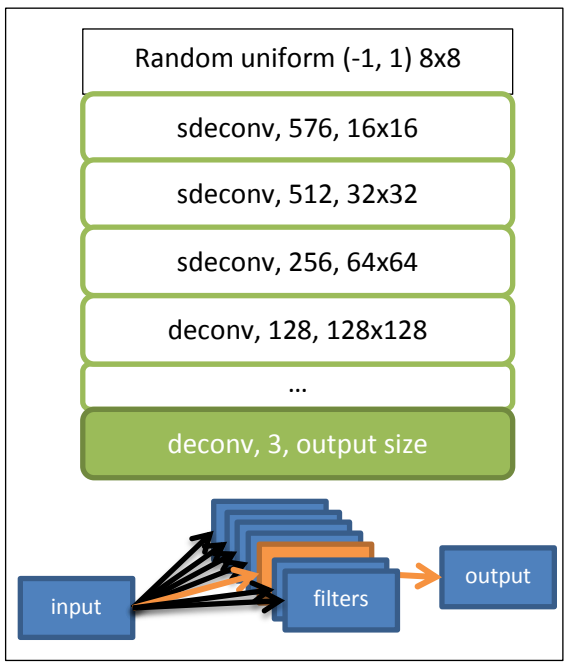

**Figure 1.** Example generator structure showing how at each 'sdeconv' (stochastic deconvolutional) layer the filters are being selected randomly.

Apart from the stochastic ensembling architecture the rest of GAN is built to standard approach with only usage of weight normalization [8] and parametric ReLU additions. For the training of the architecture the standard cGAN loss was used,

$$L_{cGAN} = \mathbb{E}_{x,y}[\log D(y)] + \mathbb{E}_{x,z}[\log(1 - D \circ G(x,z))]$$

## Experiments

Two experiments were performed to estimate the benefits of data variety with respect to the outputs of the model when using stochastic deconvolutional generator (SGAN). The comparison was made with an 'adapted GAN' described in 'On the Effects of Batch and Weight Normalization in Generative Adversarial Networks' [8] which is shown to be advantageous in regard to combating the mode collapse problem and increasing the variation within the output samples. The SGAN was constructed from the 'adapted GAN' by altering the first three generator layers. The layers were converted to stochastic deconvolutions with banks of 16 filters, which gave $16^4 = 65536$ potential combinations of filters. With the absence of a common measurement technique the comparison was performed visually. However, the final results differed significantly when compared to the normal GAN approach that we believe no extra measurements were necessary.

The first experiment used the MNIST dataset. The first notable difference was during training. It was required to restart the training of the 'adapted GAN' multiple times due to immediate collapse of the network output. This was due to wrong weight initialisation. However, when using the SGAN architecture, immediate collapse was never observed during the training process, irrespective of the initialisation.

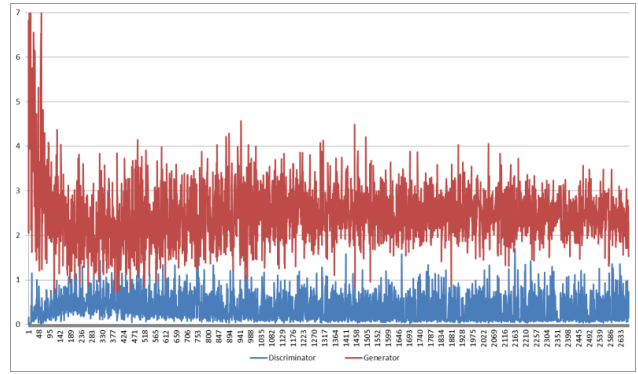

**Figure 2.** GAN loss

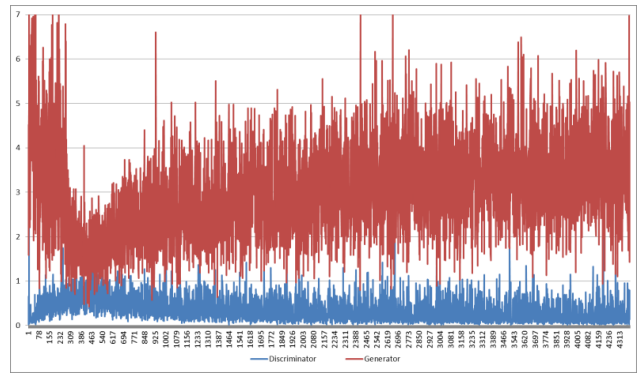

**Figure 3.** SGAN loss

The main method for comparing the output of the networks was visually inspecting the outputs. The following table shows the period output during training. The larger state of the SGAN architecture required training for a much longer period of time.

|  'Adapted GAN' Output | SGAN Output |
|---|---|

**Step 0** 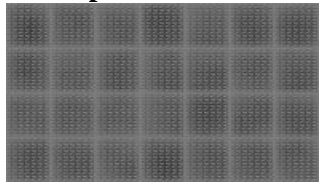

**Step 0** 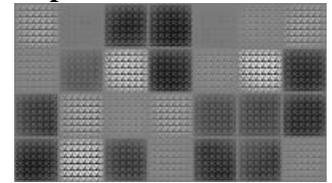

**Step 50** 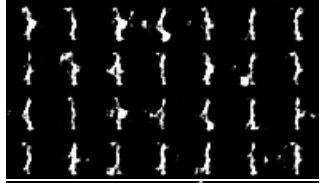

**Step 50** 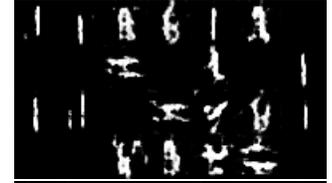

**Step 100** 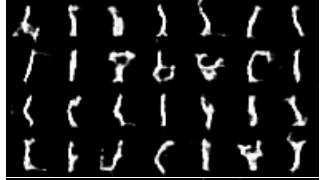

**Step 100**
Visual quality of alive paths is similar to step 100 of GAN 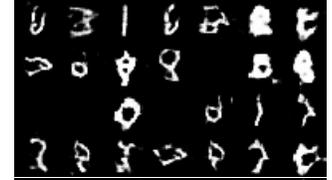

**Step 250** 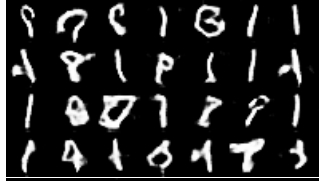

**Step 250** 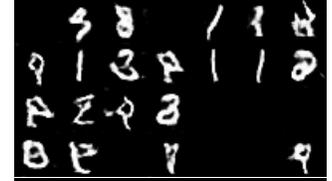

**Step 500**
Initial signs of degradation occurs; model starts to decrease variety producing only "ones" 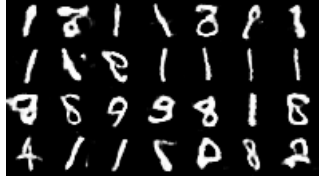

**Step 500**
Recovered dead paths, no degradation observed 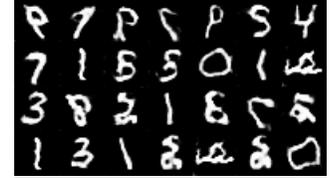

**Step 1000**
Majority of outputs are 1s 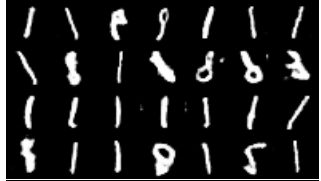

**Step 1000** 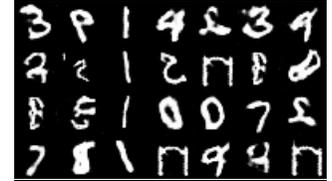

**Step 1500** 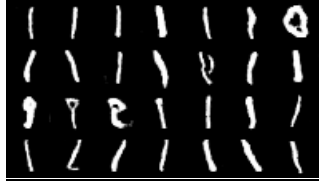

**Step 1500**
Degradation and collapse of sub-networks. Significantly better comparing to GAN 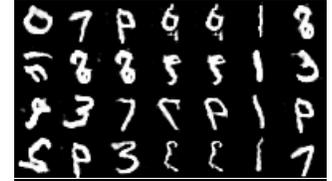

**Step 2000** 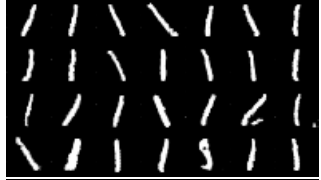

**Step 2000** 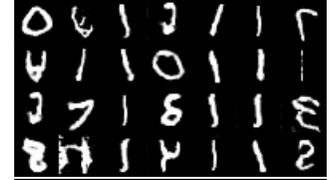

**Step 2500** 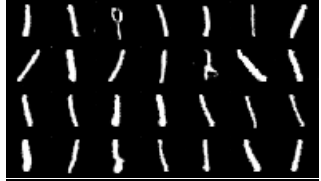

**Step 2667**
Last step of GAN model Still generating decent results with a minimal degradation 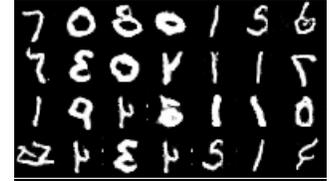

**Step 2667**
Model was not able to recover, training stopped 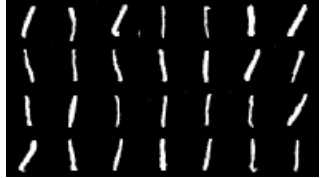

**Step 3000**
slow degradation continues, however variety is not affected comparing to quality of the output 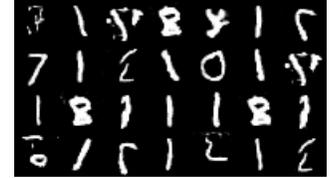

**SGAN Output (continued)**

**Step 3500**

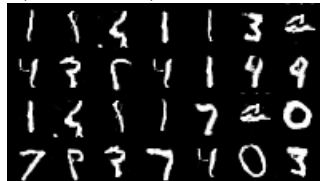

**Step 4000**

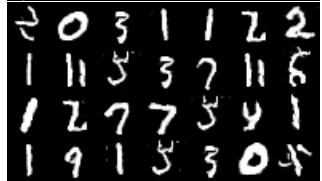

**Step 4388**

Still producing decent output. Sub-networks seem to suffer from a lack of variety, overall variety is intact.

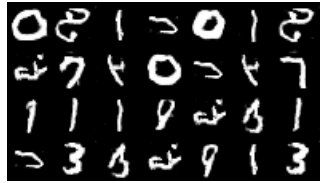

**Tommy Hilfiger Instagram account**

The SGAN architecture was also compared to the adapted GAN [8] over a dataset consisting of images scraped from the official Tommy Hilfiger Instagram account. The dataset consisted of 2350 images of all kinds of topics (Figure 4). No data argumentation was applied except random horizontal swaps. None of the known GAN architectures would be able to produce decent results based on such a small training set with such an input data variety.

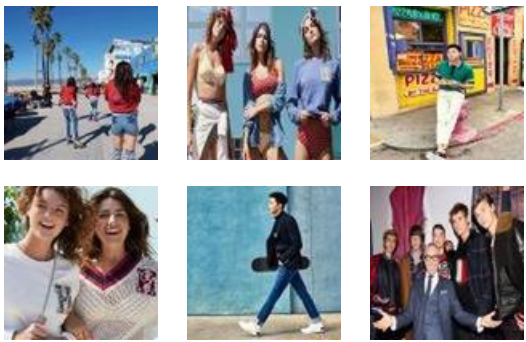

**Figure 4.** Example training images from the Tommy Hilfiger Instagram account.

The difference in stability can be seen in the progression and volatility of loss during the training of the two architectures. The SGAN is more stable (Figure 6) while the 'adapted GAN' starts to reach the practical max and 0 already (Figure 5) around step 1100.

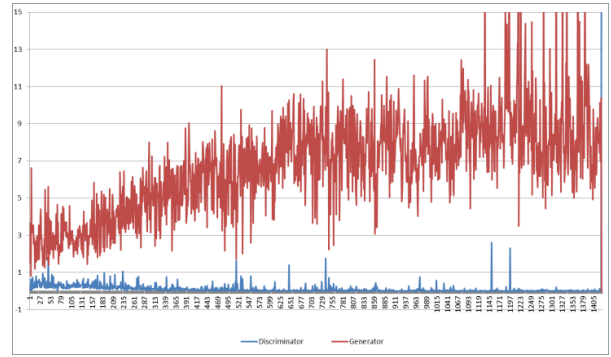

**Figure 5.** Loss for adapted GAN architecture during training.

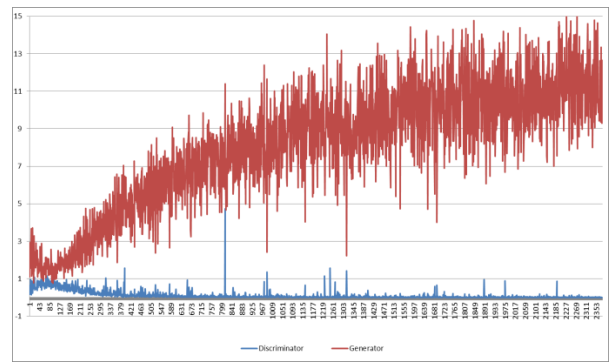

**Figure 6.** Loss for SGAN architecture during training.

| 'Adapted GAN' output | SGAN output |
|---|---|
| **Step 0** 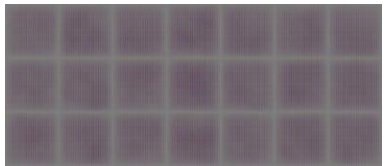 | **Step0** 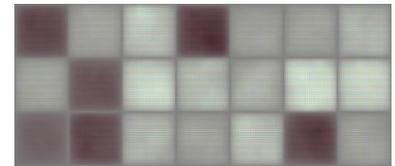 |
| **Step 50** 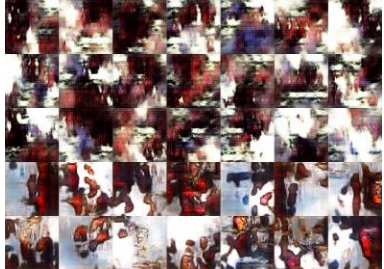 | **Step 50**<br>Greater variety observed 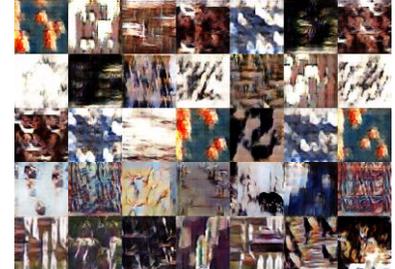 |
| **Step 100** 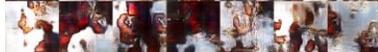 | **Step 100** 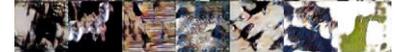 |
| **Step 500**<br>'Best' results achieved 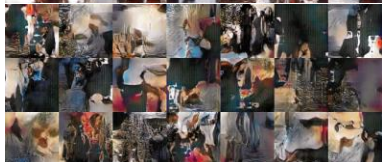 | **Step 500**<br>Some sub-networks collapsed, variety is still high 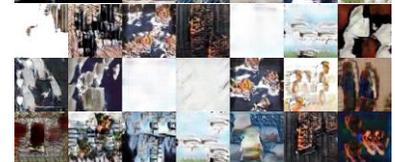 |
| **Step 654**<br>Mode collapse 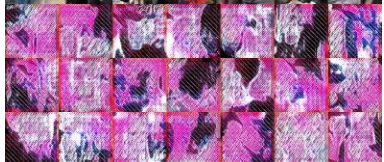 | **Step 654** 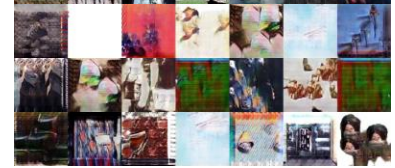 |
| **Step 700**<br>Decreased variety after recovering, started oscillating 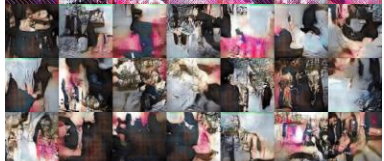 | |
| **Step 824**<br>Mode collapse again 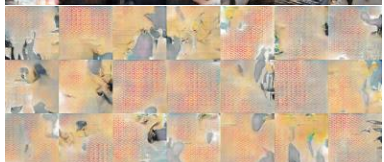 | |
| **Step 871**<br>Recovering with quality and variety degradation 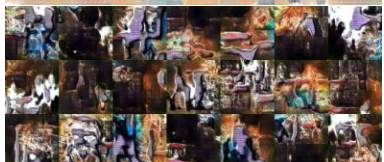 | |
| **Step 956**<br>Collapsed 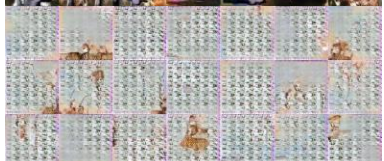 | |
| **Step 1100**,<br>Recovering with a high level of variety and quality degradation 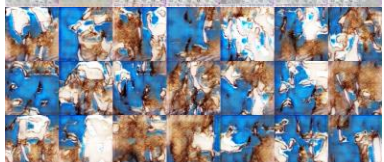 | **Step 1000**<br>Recovered collapsed paths, sub-network its own "topic" 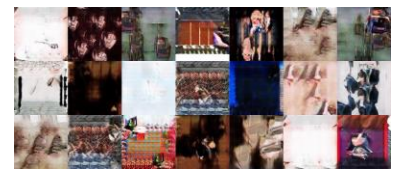 |
| **Step 1250**<br>Collapsed 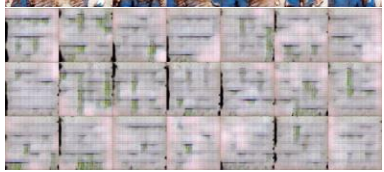 | |

**Step 1426**
Training ended, model not able to recover

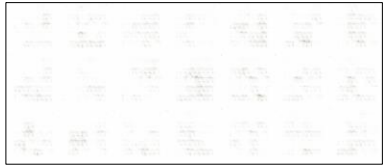

**Step 1500**
Degradation of local variety, overall variety is still present

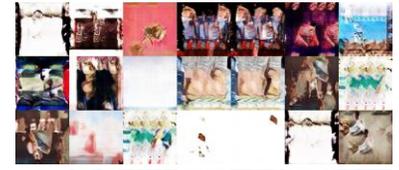

**Step 2000**
Sub-networks show signs of oscillation, overall variety still decent

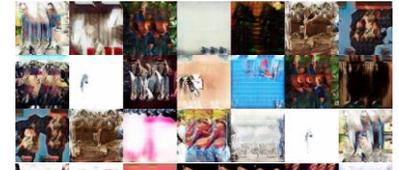

**Step 2380**
Sub-networks degradation, overall quality is superior to adapted GAN

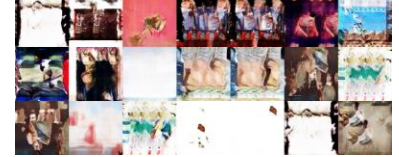

## Generative Pseudo-Adversarial Network

GAN architectures where the role of discriminator is being played by a pre-trained image classifier, can be thought of as being 'perfectly' collapsed. For any generated image, the discriminator cannot be fooled. In practice, such a network is not adversarial hence it is referred to as a Generative Pseudo-Adversarial Network (GPAN).

Following the classic approach, such a model can never be trained and will become stuck at the very start with the theoretically infinite gradients of the generator. Suggested below is a model that behaves similarly to a typical adversarial model by applying stochastic ensembling and the 'adapted PatchGAN' described in 'Another way of restoring distribution of an image classifier' [9]. The suggested model follows the 'adapted PatchGAN', with minor modifications, applying stochastic ensembling on deep levels of the generator (Figure 7) and extending the output to a higher resolution of 1024.

In this architecture a stochastic deconvolution layer and convolutional layers are constructed using filters of size 4, applied with a stride of size 2. PReLU (Parameterised Leaky ReLU function initialised to 0.2) was applied to improve model fitting [7] unless it was the last layer for which tanh function was used. Weight normalization was also used, so as to improve stability [8].

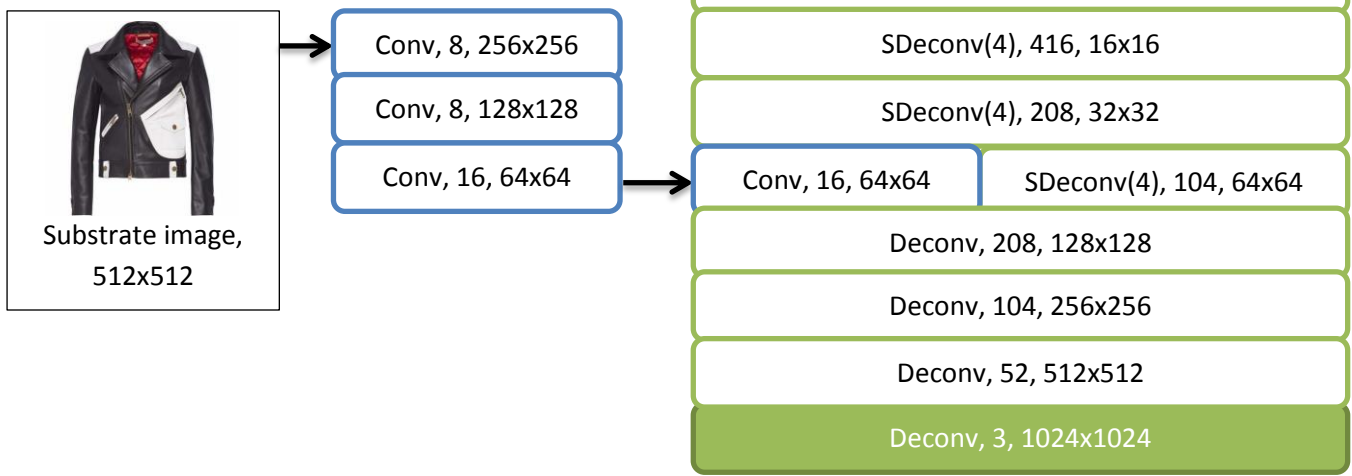

**Figure 7.** Generator part of GPAN architecture featuring 5 SDeconv layers

In the construction of the GPAN, the generator part (Figure 7) used 5 stochastic deconvolution layers with filter banks consisting of 4 sets of filters. This gives a theoretical maximum variety of 1024 combinations. The discriminator remained unchanged, but received a resized, 512 × 512, copy of the output produced by generator.

A few additions were made to the architecture to achieve better quality of the output. The importance regions, in accordance with the grad-cam methodology [10], of the pool5 layer in the VGG16 classifier were extracted and masked in a conditional GAN (cGAN) loss. The purpose of this was to allow the output to satisfy either the VGG or discriminator losses independently. In addition, to improve the output variety in each batch of images, $B$ the constraint to contain only one set of wanted categories from the classifier was applied.

The same losses for the 'adapted PatchGAN' were used. For the cGAN part of the network,

$$L_{cGAN} = \mathbb{E}_{x,y}[\log D(x,y)] + \mathbb{E}_{x,z}\left[\log\left(1 - D(x, G(x,z))\right)\right]$$

The masking loss was applied to keep the background of the image as white as possible,

$L_m = 1 - M(G)$ where $M(G)$ is the masked output of the generator. The substrate loss to influence the model to generate an image $G$ as similar to the substrate $T$ as possible,

$$L_{sub} = |T - G| - \log\left(1 - \left|\frac{T-G}{2}\right|^2\right)$$

As well as the same loss for the output activating the VGG classifier,

$$L_{VGG} = L_p + L_n$$

where $L_p$ is the loss for required classes not appearing in the output and $L_n$ is the loss associated with classes being identified that were not required. [9].

Let, $M_i(x)$, $\forall i \in M = \{4, 8, 16, 32, 64\}$ be the feature map from each sdeconv layer, with shape $i \times i \times$ depth. Each feature map was compared between images in the same batch to construct a 'Split Loss'. The purpose of this loss was to influence the network to develop different feature maps for each image in batch. Defining $F: \mathbb{R}^3 \to \mathbb{R}^3$ as the function which measures the element-wise difference between two 3-dimensional feature maps,

$$F(a,b) = \frac{1}{|x|}\sum_x \tanh\left(\left(\frac{a_x - b_x}{25}\right)^4\right)$$

For training with a batch $B$ the split loss is defined as,

$$L_{split} = \sum_{i \in M} \frac{2w_i}{|B|\cdot(|B|-1)} \sum_{k,j \in B} \log\left(F(M_i(k), M_i(j))\right)$$

The term $w_i$ refers to a scaling factor dependant on the layer $i$:

$$w_4 = 0.03125, \quad w_8 = 0.0625$$

$$w_{16} = 0.125, \quad w_{32} = 0.25$$

$$w_{64} = 0.53125$$

The combination of losses that was observed to be the most beneficial was

$$L = 3L_{split} + 10L_{cGAN} + 100L_{VGG} + 25L_m + 100L_{sub}$$

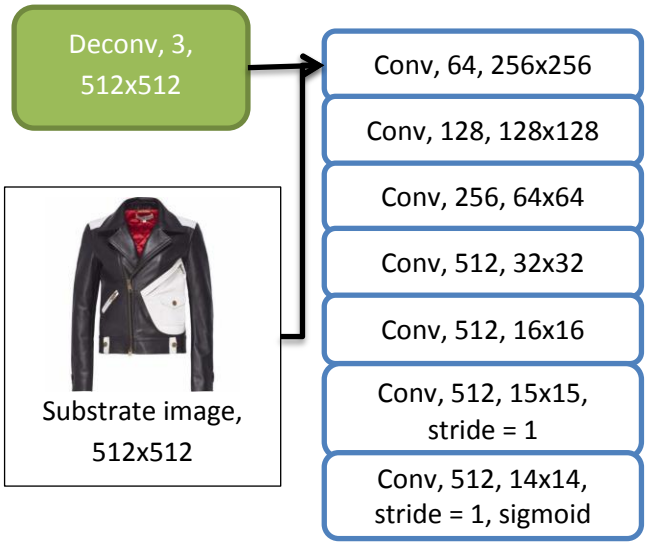

**Figure 8.** Discriminator part of GPAN architecture

## Results

Significant variety in the output data was achieved giving an impression of dynamic patterns generation as if produced by a classic generative adversarial network. At the each iteration, 1 out of 1024 paths was chosen giving the appearance of random behaviour. As a consequence even if the same target group is selected a PGAN generates different images at each run.

In the following examples, the images were generated for each target category. Different paths were used for each image, resulting in the different patterns.

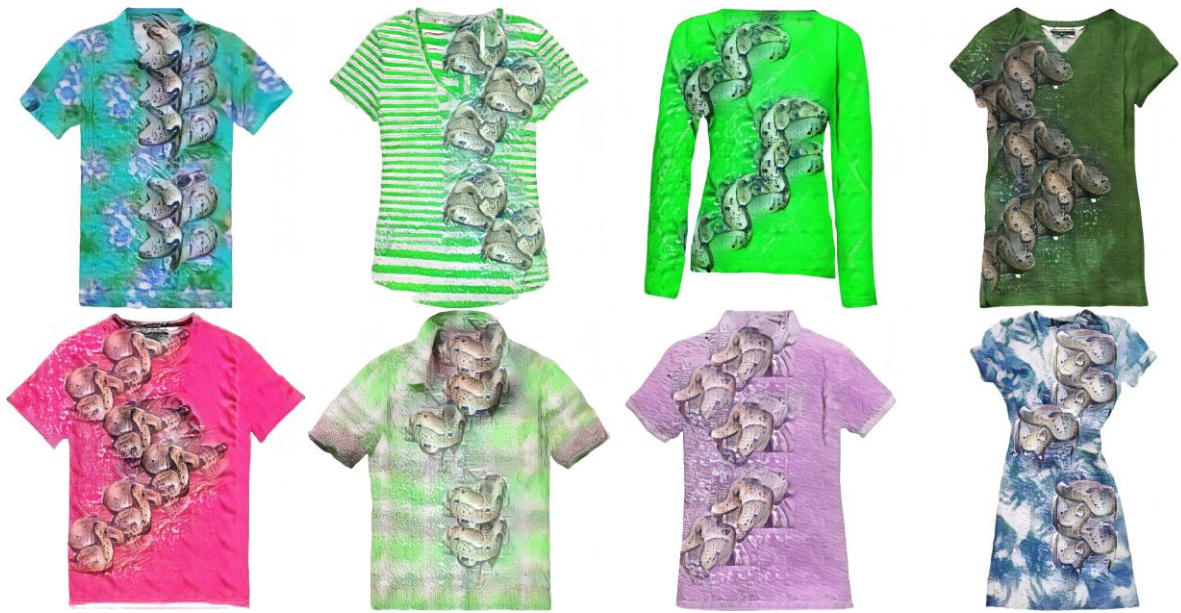

Python

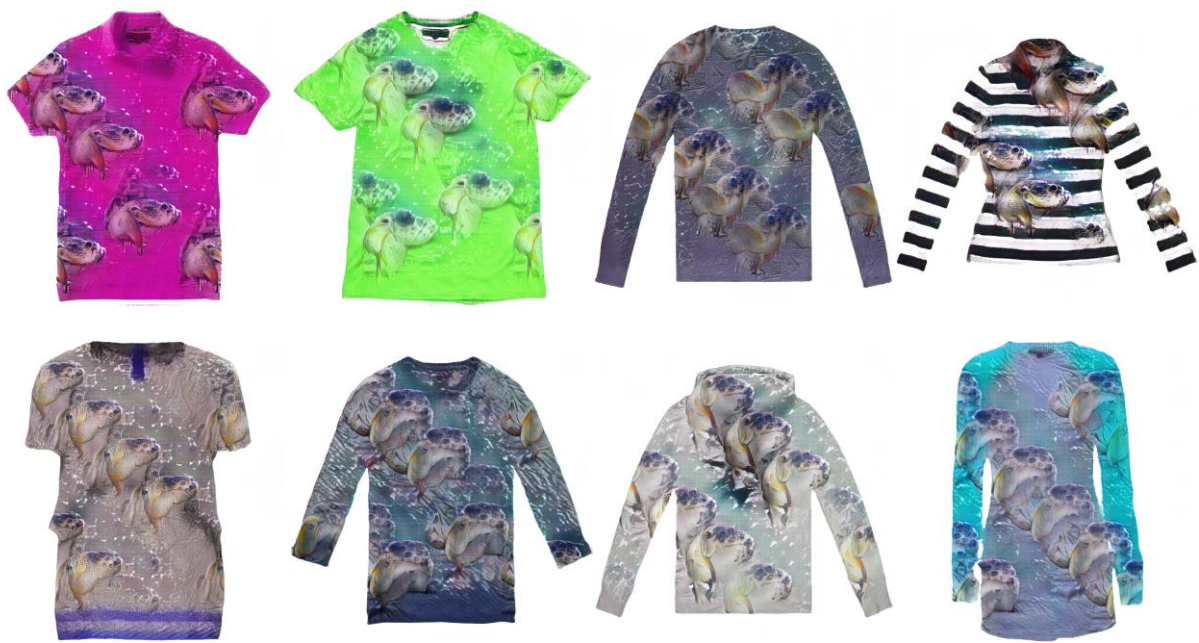

Turtle

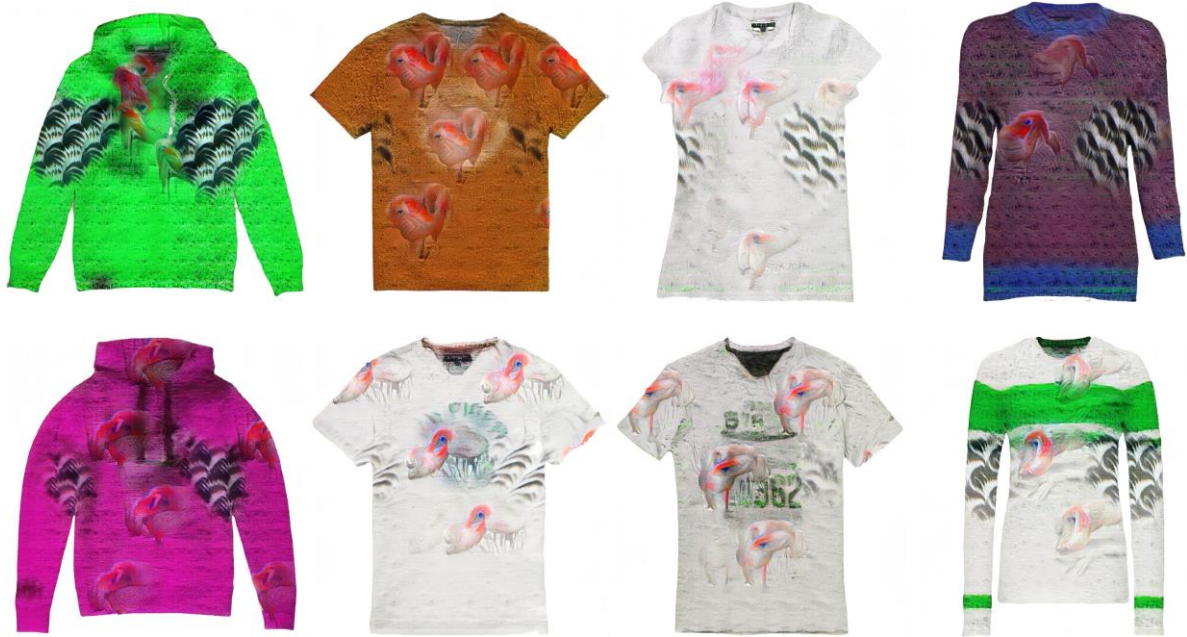

Zebra and flamingo

## Conclusion

Stochastic generative models can be beneficial for a wide range of applications where generative transformations have unlimited solutions. In such instances the suggested approach can help in covering a subset without making the model's structure significantly bigger and/or complicated. In particular it can be useful for chatbot applications. Answers to the same question could be formulated in a different way giving the appearance of more human speech.